\documentclass[letterpaper, 10 pt, journal, twoside]{IEEEtran}
\usepackage{relsize}
\usepackage{xspace}
\usepackage{color}
\usepackage{graphics}
\usepackage{graphicx}

\usepackage{amsmath}

\usepackage{amsfonts}
\usepackage{amssymb}
\usepackage{amsthm}

\usepackage{float}
\usepackage{url}
\usepackage{booktabs}
\usepackage{multirow}
\usepackage{hhline}
\usepackage[capitalise]{cleveref}
\usepackage{dsfont}
\usepackage{cancel}

\usepackage{subcaption}
\usepackage{catchfilebetweentags}

\usepackage{booktabs}

\usepackage{csvsimple}

\usepackage{tikz}
\usetikzlibrary{positioning}

\newtheorem{problem}{Problem}

\newcommand{\acronym}[1]{{#1}\xspace}

\newcommand{\shortname}{\acronym{Swarm-SLAM}}

\usepackage{bm}

\newcommand{\bdmath}{\begin{dmath}}
\newcommand{\edmath}{\end{dmath}}
\newcommand{\beq}{\begin{equation}}
\newcommand{\eeq}{\end{equation}}
\newcommand{\bdm}{\begin{displaymath}}
\newcommand{\edm}{\end{displaymath}}
\newcommand{\bea}{\begin{eqnarray}}
\newcommand{\eea}{\end{eqnarray}}
\newcommand{\beal}{\beq \begin{array}{ll}}
\newcommand{\eeal}{\end{array} \eeq}
\newcommand{\beas}{\begin{eqnarray*}}
\newcommand{\eeas}{\end{eqnarray*}}
\newcommand{\ba}{\begin{array}}
\newcommand{\ea}{\end{array}}
\newcommand{\bit}{\begin{itemize}}
\newcommand{\eit}{\end{itemize}}
\newcommand{\ben}{\begin{enumerate}}
\newcommand{\een}{\end{enumerate}}

\newcommand{\hide}[1]{}

\newcommand{\hiddenText}{{\color{gray} hidden text.}}
\newcommand{\hideWithText}[1]{\hiddenText}

\newcommand{\blue}[1]{{\color{blue}#1}}

\newcommand{\linkToPdf}[1]{\href{#1}{\blue{(pdf)}}}
\newcommand{\linkToPpt}[1]{\href{#1}{\blue{(ppt)}}}
\newcommand{\linkToCode}[1]{\href{#1}{\blue{(code)}}}
\newcommand{\linkToWeb}[1]{\href{#1}{\blue{(web)}}}
\newcommand{\linkToVideo}[1]{\href{#1}{\blue{(video)}}}
\newcommand{\award}[1]{\xspace} %

\newcommand{\localedges}{\mathcal{E}^{\text{local}}}
\newcommand{\fixededges}{\mathcal{E}_{\text{fixed}}^{\text{global}}}
\newcommand{\candidateedges}{\mathcal{E}_{\text{candidate}}^{\text{global}}}

\begin{document}

\title{\shortname: Sparse Decentralized Collaborative Simultaneous Localization and Mapping Framework for Multi-Robot Systems}

\author{Pierre-Yves Lajoie, Giovanni Beltrame%
\thanks{ 
  This work was partially supported by a Vanier Canada Graduate Scholarships
  Award and by the Canadian Space Agency.
 }%
\thanks{Department\,of\,Computer\,and\,Software\,Engineering, \mbox{Polytechnique Montr\'eal}, Montreal, Canada, \newline
{\tt\scriptsize\,\{pierre-yves.lajoie, giovanni.beltrame\}@polymtl.ca}}
}

\maketitle

\begin{abstract}
Collaborative Simultaneous Localization And Mapping (C-SLAM) is a vital
component for successful multi-robot operations in environments without an
external positioning system, such as indoors, underground or underwater. In this
paper, we introduce \shortname, an open-source C-SLAM system that is designed to
be scalable, flexible, decentralized, and sparse, which are all key properties
in swarm robotics. Our system supports lidar, stereo, and RGB-D
sensing, and it includes a novel inter-robot loop closure prioritization
technique that reduces communication and accelerates convergence. We evaluated
our ROS~2 implementation on five different datasets, and in a real-world
experiment with three robots communicating through an ad-hoc network. Our code
is publicly available: \url{https://github.com/MISTLab/Swarm-SLAM} \end{abstract}

\begin{IEEEkeywords}
SLAM, Multi-Robot Systems, Collaborative Perception, Swarm Intelligence
\end{IEEEkeywords}

\section{Introduction}\label{sec:intro}

\IEEEPARstart{C}{ollaborative} perception is an important problem for the future
of robotics. The shared understanding of the environment it provides is a
prerequisite to many applications from autonomous warehouse management to
subterranean exploration. One of the most powerful tools for robotic perception
is Simultaneous Localization And Mapping (SLAM) which tightly couples the
geometric perception of the environment with state
estimation~\cite{cadenapresentfuturesimultaneous2016}. In addition to producing
high-quality maps of the robot surroundings, it provides localization estimates
that are essential for planning and control. However, single-robot SLAM
estimates are local in the individual robot reference frame. Therefore, when
multiple robots operate in GPS-denied environments, they do not share
situational awareness unless they manage to connect, or merge, their local maps.
To solve this problem, Collaborative SLAM (C-SLAM) searches for inter-robot map
links and uses them to combine the local maps into a shared global understanding
of the environment.
One of the main practical challenges in C-SLAM is resource
management~\cite{lajoiecollaborativesimultaneouslocalization2022a}, in particular
considering the severe communication and computation limitations of mobile
robots. Those limitations need to be addressed to achieve real-time performance,
especially when a large number of robots work together. While effective in
some scenarios, centralized C-SLAM solutions, which rely on a single server for
data association and optimization, suffer from a communication bottleneck between the robots and the
server, which limits their scalability. Besides, due to networking coverage
challenges in large indoor or subterranean environments, robots
cannot realistically maintain a stable connection to a central server. Thus,
decentralized solution relying only on occasional communication
between the robots are better suited for large-scale deployment.
\ExecuteMetaData[explanation_figures.tex]{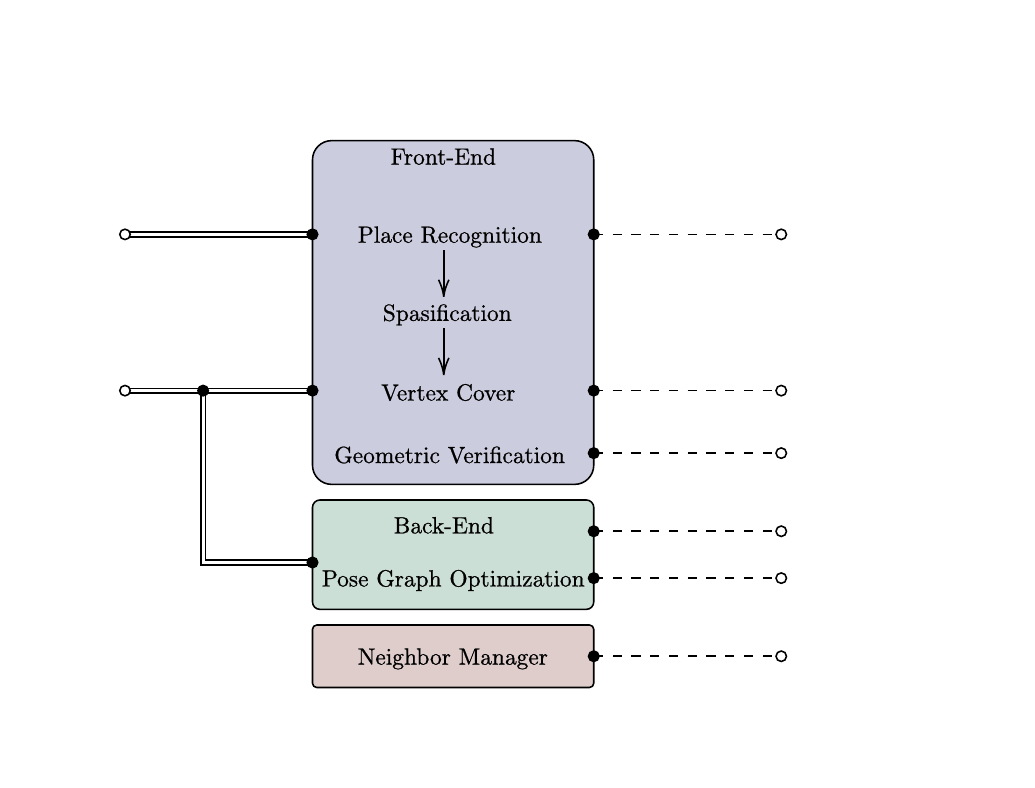}
While collaborative perception within small teams of autonomous robots is
currently challenging, we believe it useful to look forward to very large teams,
or swarms of robots and start tackling the problems specific to this scale of
deployment. Prior works on swarm robotics have identified a few key properties
required for swarm compatibility~\cite{brambillaswarmroboticsreview2013}
such as: communication and sensing must be local to the
robot neighborhood, and robots should not rely on a centralized
authority or global knowledge. In the specific case of
C-SLAM~\cite{kegeleirsswarmslamchallenges2021}, we consider the following four properties described in \cref{sec:system_overview,sec:frontend}: scalability, flexibility, decentralization, and sparsity.

In this paper, we propose novel techniques assembled in a complete
resource-efficient C-SLAM framework compliant with these key swarm compatibility
properties. Our approach is fully decentralized, supports different types of
sensors (stereo cameras, RGB-D cameras, and lidars), 
and requires significantly less communication than previous
techniques. To reduce data exchanges, we introduce a novel budgeted
approach to select candidate inter-robot loop closures based on algebraic
connectivity maximization, inspired from recent work on pose graph
sparsification~\cite{dohertyspectralmeasurementsparsification2022}. This
preprocessing of place recognition matches allows us to achieve accurate C-SLAM
estimates faster and using fewer communication resources. Moreover, we leverage
 advances in robotic
software engineering, to make our framework compatible with ad-hoc networks.
In summary, we offer the following \textbf{contributions}:
\begin{itemize}
\item A sparse budgeted inter-robot loop closure detection algorithm under
  communication constraints based on algebraic connectivity maximization;
\item A decentralized approach to neighbor management and pose graph
  optimization suited for sporadic inter-robot communication;
\item A swarm-compatible open-source framework which supports
  lidars, as well as stereo or RGB-D cameras;
\end{itemize}
We extensively evaluate of the overall system performance on datasets and in a
real-world experiment.
\section{Background and Related Work}\label{sec:related_work}

\subsection{Collaborative SLAM}

C-SLAM systems can usually be divided into two categories: centralized and
decentralized. Centralized systems rely on a remote base station to aggregate map data and compute the
global SLAM estimates for all the robots. 
However, in those systems, the robots need a reliable permanent connection with
the base station, and the scalability is severely limited by the communication
bottleneck to the central server. Such stringent
networking constraints are often unrealistic, especially in large environments.
Decentralized approaches, relying only on occasional communication links between
robots and without any need for a central authority, are preferred in those
scenarios. However, decentralized systems are limited by the onboard
computation and communication capabilities of the robots, and they require more
sophisticated data management and bookkeeping strategies to obtain accurate SLAM
estimates~\cite{lajoiecollaborativesimultaneouslocalization2022a}.
Similar to single robot SLAM systems, C-SLAM contains two parts commonly named
front-end and back-end, see \cref{fig:overview}. The front-end is in charge of feature extraction and
data association, while the back-end performs state
estimation~\cite{cadenapresentfuturesimultaneous2016}.

\subsubsection{Front-End}

The most challenging step in the C-SLAM front-end is the detection and computation of
inter-robot loop closures in a resource-efficient manner. Inter-robot loop closures correspond to common
features or places previously visited by two or more robots. Those shared features
between the robots maps act as stitching points to merge the local
maps together and obtain a shared (global) reference frame. 
Since the communication cost of sharing entire maps is usually prohibitive,
 inter-robot loop closure detection can be performed in two
stages~\cite{cieslewskidataefficientdecentralizedvisual2018,lajoiedoorslamdistributedonline2020}.
In the first stage, compact global descriptors of
images~\cite{bertonrethinkingvisualgeolocalization2022a} or lidar
scans~\cite{kimscancontextegocentric2018}, are shared between the robots for
place recognition. Similarity scores are computed between the global descriptors
from both robots to recognize places, or overlaps, between their respective
maps. The recognized places then correspond to loop closure candidates for the
second stage. 
In the second stage, for each candidate with high global
descriptors similarity, the corresponding costly local descriptors such as 3D keypoints
or scans are transmitted to compute the geometric registration between
the two robots images or scans. 

\subsubsection{Back-End}

The role of the C-SLAM back-end is to estimate the most likely poses and map
from the noisy measurements gathered by all robots. To this end, Choudary et
al.~\cite{choudharydistributedmappingprivacy2017} propose the distributed
Gauss-Seidel (DGS) technique which allows robots to converge to a globally
consistent local pose graph by communicating only the pose estimates involved in
inter-robot loop closures, and therefore preserving the privacy of their whole
trajectories. Tian et al.~\cite{tiandistributedcertifiablycorrect2021}
significantly improve on that approach and provide a certifiably correct
distributed solver for pose graph optimization. This technique performs multiple
exchanges between the robots until they converge to globally consistent local
solutions. 
In a different
vein, recent work by Murai et al.~\cite{murairobotwebdistributed2022} laid the
foundation for larger-scale multi-robot collaborative localization based on
Gaussian Belief Propagation.
One of the main challenges in both single-robot and collaborative SLAM is the
frequent occurrence of erroneous measurements among inter-robot loop closures due
to perceptual aliasing~\cite{lajoiemodelingperceptualaliasing2019}. While many
techniques exist for the single-robot problem,
Lajoie et al.~\cite{lajoiedoorslamdistributedonline2020} first combined DGS with Pairwise
Consistency Maximization
(PCM)~\cite{mangelsonpairwiseconsistentmeasurement2018}, which computes the
maximal clique of pairwise consistent inter-robot measurements, to perform
robust and distributed optimization. More recently,
Yang et al.~\cite{yanggraduatednonconvexityrobust2020} introduced the Graduated
Non-Convexity (GNC) algorithm, a general approach for robust estimation
on various problems including pose graph optimization. GNC was integrated
with~\cite{tiandistributedcertifiablycorrect2021} in a robust distributed
solver (D-GNC)~\cite{tiankimeramultirobustdistributed2022}.

\ExecuteMetaData[explanation_figures.tex]{frameworks}

\subsubsection{Open-Source C-SLAM Systems}

Many open-source C-SLAM systems have been proposed in the recent years.
Cieslewski et al.~\cite{cieslewskidataefficientdecentralizedvisual2018}
introduce DSLAM, which uses CNN-based global descriptors for distributed place
recognition, and DGS for estimation.
DOOR-SLAM~\cite{lajoiedoorslamdistributedonline2020} robustified the approach by
integrating PCM for outlier rejection and adapted it for sporadic inter-robot
communication. DiSCo-SLAM~\cite{huangdiscoslamdistributedscan2022} extends those
ideas to lidar-based C-SLAM using ScanContext global
descriptors~\cite{kimscancontextegocentric2018}.
Kimera-Multi~\cite{tiankimeramultirobustdistributed2022} integrates D-GNC and
incorporates semantic data in the resulting maps.
In an other line of work, centralized C-SLAM system have also evolved
considerably. 
The lidar-based system
LAMP~2.0~\cite{changlamprobustmultirobot2022a} introduces a centralized Graph
Neural Network-based prioritization mechanism to predict the outcome of pose
graph optimization for each inter-robot loop closure candidates. The multi-modal
maplab~2.0~\cite{cramariucmaplabmodularmultimodal2022} supports heterogeneous
groups of robots with different sensor setups.
In contrast, \shortname combines the latest advances from previous frameworks
and introduce a new sparse inter-robot loop closure prioritization to further reduce
communication. Additionally, unlike previous techniques, \shortname leverages
 ROS~2~\cite{macenskirobotoperatingsystem2022} and introduces a neighbor management system to seamlessly integrate C-SLAM
 with ad-hoc networking.
\cref{tab:frameworks} offers a
comparison of the various systems based on key desirable properties.
We refer the reader
to~\cite{lajoiecollaborativesimultaneouslocalization2022a} for a thorough survey
on C-SLAM.

\subsection{Graph Sparsification for C-SLAM}

The ever-growing map and pose graph during long-term operations is an important
memory and computation efficiency challenge in both single-robot and multi-robot
SLAM. One favored solution is graph
sparsification, which aims to approximate the complete graph with as few edges
as possible, mainly by removing redundant edges that are not providing new
information during the estimation process. To this end, Doherty et
al.~\cite{dohertyspectralmeasurementsparsification2022} formulate the graph
sparsification of single robot pose graphs as a \textit{maximum algebraic
  connectivity augmentation} problem, and solve it efficiently using a more
tractable convex relaxation.
In this paper, instead of sparsifying the pose graph after all the measurements
have already been computed, we aim to preemptively sparsify the inter-robot loop
closure candidates generated by the place recognition module. This way, we can
prioritize the geometric verification of inter-robot loop closures that will
approximate the full pose graph, thus avoiding wasting resources on redundant
measurements. Importantly, unlike other work maximizing the determinant of the
information matrix~\cite{tianresourceawareapproachcollaborative2021}, we
leverage the results from~\cite{dohertyspectralmeasurementsparsification2022}
and focus on the algebraic connectivity of the pose graph which has been shown
to be a key measure of estimation
accuracy~\cite{dohertyperformanceguaranteesspectral2022}
(i.e., higher algebraic connectivity is associated with lower estimation error).
Solving a similar problem, Denniston et
al.~\cite{dennistonloopclosureprioritization2022a} prioritize loop closure
candidates based on point cloud characteristics, the proximity of known beacons,
and the information gain predicted with a graph neural network. Interestingly,
Tian et al.~\cite{tianspectralsparsificationcommunicationefficient2022} explore spectral
sparsification in the C-SLAM back-end to reduce the required communication
during distributed pose graph optimization.
\section{System Overview}\label{sec:system_overview}

As described in \cref{fig:overview}, \shortname is composed of three modules.
First, to enable decentralization, the neighbor management module continuously
tracks which robots are in communication range (i.e., neighbors that can be
reached reliably) and what data has been exchanged. Robots publish heartbeat
messages at a fixed rate such that network connectivity can be evaluated
periodically. To make the system scalable (see \cref{prop:scal}), the other
modules query the neighbor management process to determine which robots, if any,
are available, and orchestrate the operations.
\ExecuteMetaData[properties.tex]{scalability}

The front-end takes as input odometry estimates (obtained using an arbitrary
technique) along with synchronized sensor data (see \cref{prop:flex}). Upon
reception, the front-end extracts global (e.g. compact learned representation) and local
descriptors (e.g. 3D keypoints). Global descriptors allow us to identify
candidate place recognition matches (i.e., loop closures) between the robots,
then local descriptors are used for 3D registration.

\ExecuteMetaData[properties.tex]{flexibility}

\ExecuteMetaData[experimentation_figures.tex]{viz-spectral}

In our decentralized (see \cref{prop:dec}) back-end, the resulting intra-robot
and inter-robot loop closure measurements are combined with the odometry
measurements into a pose graph. Local pose graphs are transmitted to the robot
selected, through neighbor management negotiation, to perform the optimization
and the resulting estimates are sent back to the respective robots.
\ExecuteMetaData[properties.tex]{decentralized} Current pose estimates,
resulting from the whole process, are made available periodically in the form of
ROS~2 messages for a minimally invasive integration into existing robotic
systems. To avoid needless bandwidth use, the neighbor manager keeps track of
which measurements have been exchanged. Mapping data for planning or
visualization can also be queried at the cost of additional computation and
communication. For debugging purposes, we provide a minimal visualization tool
which opportunistically collects mapping data from robots in communication
range. Overall, we divided place recognition, geometric verification and pose
graph optimization into modular and decoupled processes with clear data
interfaces to enable researchers to leverage Swarm-SLAM to easily test new ideas
in each subsystems.

\section{Front-End}\label{sec:frontend}

Similar to many comparable inter-robot loop closure
detection techniques
(e.g.~\cite{cieslewskidataefficientdecentralizedvisual2018,lajoiedoorslamdistributedonline2020,tiankimeramultirobustdistributed2022}),
we adopt a two stage approach in which global matching generate candidate place
recognition matches that are verified using local features in the latter stage, i.e. local matching.

\subsection{Global Matching}

For each keyframe, compact descriptors, that can be compared with a similarity
score, are extracted from sensor data and broadcast to neighboring robots.
When two robots meet, we perform bookkeeping to determine which global
descriptors are already known by the other robot and which ones need to be
transmitted. We use ScanContext~\cite{kimscancontextegocentric2018} as global
descriptors of lidar scans and the recent CNN-based
CosPlace~\cite{bertonrethinkingvisualgeolocalization2022a} for images. 
We use
nearest neighbors based on cosine similarity for descriptor matching. 
Once matches are computed, \shortname offers two candidate prioritization
mechanisms: a \textbf{greedy} prioritization algorithm, used in prior
work~\cite{cieslewskidataefficientdecentralizedvisual2018,lajoiedoorslamdistributedonline2020,tiankimeramultirobustdistributed2022,huangdiscoslamdistributedscan2022},
and a novel \textbf{spectral} approach.
To perfom the candidate prioritization, we define the multi-robot pose graph as:
\begin{align}
  \mathcal{G} &= (V,\localedges,\mathcal{E}^{\text{global}})\\
  V &= (V_{1}, \dots, V_n)\\
  \localedges &= (\localedges_{1}, \dots, \localedges_{n})\\
  \mathcal{E}^{\text{global}} &= (\fixededges, \candidateedges)
\end{align}
where $V$ are the vertices from every $n$ robots pose graphs, each vertex
corresponding to a keyframe; $\localedges$ are the local pose graphs edges such
as odometry measurements and intra-robot loop closures; and
$\mathcal{E}^{\text{global}}$ are the global pose graph edges corresponding to
inter-robot loop closures. $\mathcal{E}^{\text{global}}$ is further divided
between $\fixededges$ which contains the fixed measurements that have already
been computed, and the candidate inter-robot loop closures $\candidateedges$ on
which the prioritization is performed. Detailed measurements (i.e., pose
estimates) are not required for our proposed candidate prioritization mechanism.
Therefore, fixed measurements, both local and global, are undirected unweighted
edges between two vertices, and candidates edges contain an additional weight
value corresponding to their respective similarity score. This reduced
multi-robot pose graph can be built directly from the global matching
information and does not require any additional inter-robot communication.

The number of edges $B$ to select at each time step is set by the user. This
budget should reflect the communication and computation capacities of the
robots. The common candidate prioritization approach widely used prior works is
a basic \textit{greedy prioritization} in which the top $B$ candidates with the
highest similarity scores are selected.

In our proposed \textit{spectral prioritization} process, we frame pose graph sparsification as a candidate
prioritization problem, and leverage recent work on spectral sparsification. 
We observe that the two problems are mathematically equivalent, one being solved before loop closure computation and the other after.
Specifically, we perform sparsification on the candidate
inter-robot matches before computing the corresponding 3D measurements, reducing
resource usage for the costly inter-robot geometric verification of redundant
candidates, and achieving better accuracy (see \cref{prop:sparse}).

\ExecuteMetaData[properties.tex]{sparsity}

As shown in~\cite{dohertyperformanceguaranteesspectral2022}, the algebraic
connectivity of the pose graph controls the worst-case error of the solutions of
the SLAM \textit{Maximum Likelihood Estimation} problem. The
pose graph algebraic connectivity corresponds to the second-smallest eigenvalue
$\lambda_2$ of the \textit{rotation weighted Laplacian} with entries for pairs
of vertices $(i,j)$ defined as:
\begin{align}
  L_{ij} = \begin{cases} \sum_{(i,j') \in \delta(i)} \kappa_{ij'},& i = j, \\
    -\kappa_{ij},& \{i,j\} \in \mathcal{E}, \\
    0,& \{i,j\} \notin \mathcal{E}. \end{cases}
  \label{eq:laprot}
\end{align}
where $\kappa_{ij}$ denotes the edge weight and $\delta(i)$ is the set of edges
incident to vertex $i$. 
Instead of using a noise model for the edge weights as in~\cite{dohertyspectralmeasurementsparsification2022}, we use the similarity score $s_e \in [0,1]$ from global matching as confidence metric. 
Thus, we define $\kappa_{ij}=1 \; \forall
\; e \in (\localedges,\fixededges)$, and $\kappa_{ij}=s_e \; \forall \; e \in
\candidateedges$. 
This approach
forgoes the need to communicate additional information regarding the edges'
estimated noise level.
However, it is important to note that it loses the theoretical guarantees from~\cite{dohertyspectralmeasurementsparsification2022}, yet we show in~\cref{sec:experiments} that this heuristic approach works well in realistic cases.

For our purposes, we leverage the property that the Laplacian $L$ can be
expressed as the sum of subgraph Laplacians corresponding to each of its edges
to define the augmented pose graph Laplacian as follows:
\begin{align}
  L(\omega) \triangleq L^{\localedges} + L^{\fixededges} + \sum_{e \in \candidateedges}\omega_e L_e
\end{align}
where $\omega_e \in \{0,1\}$ is the binary variable which determines the
prioritization of candidate edge $e$.
Therefore, according to our previously stated goal, we aim to select the subset
$\mathcal{E}^\star \subseteq \candidateedges$ of fixed budgeted size $|
\mathcal{E}^{\star} | = B$ which maximizes the algebraic connectivity $\lambda_2
(L(\omega))$:
\begin{problem}{Candidate prioritization via Algebraic Connectivity Maximization}
  \begin{align}
    \begin{gathered}
      \max_{\omega_e \in \{0,1\}}  \lambda_2 (L(\omega))\\
      |\omega| = B.
    \end{gathered}
  \end{align}
  \label{prob:acm}
\end{problem}

\cref{prob:acm} is NP-Hard~\cite{mosk-aoyamamaximumalgebraicconnectivity2008}
due to the integrality constraint on $\omega_e$. Therefore, we relax the integrality
constraints and, when necessary, we round the optimization result to the nearest
solution in the feasible set of~\cref{prob:acm}. We solve the relaxed problem
using the simple and computationally inexpensive approach developed
in~\cite{dohertyspectralmeasurementsparsification2022}. It is important to note
that this approach requires the pose graph to be connected, so we first perform
greedy prioritization up until at least one inter-robot loop closure exists
between the local pose graphs. We also use the greedy solution as initial guess
for the algebraic connectivity maximization.

In \cref{fig:viz-spectral}, we present a visualization of our spectral approach results
in comparison to the ones obtained with the standard greedy approach. We can see
that the candidates selected using our spectral technique are more evenly
distributed along the pose graph while the greedy candidates are mostly
concentrated in high-similarity areas. Our selected candidates are therefore
less redundant for the estimation process.

\subsection{Local Matching}
Once the inter-robot loop closure candidates are selected, the next step is to
perform local matching (i.e., geometric verification). This step leverages
larger collections of local features, keypoints or point clouds depending on the sensor, to compute the 3D
relative pose measurement between the candidate's two vertices. 
To avoid computing the same loop closure twice and to reduce the communication burden of geometric verification,
we follow~\cite{giamoutalkresourceefficientlyme2018} and formulate the vertices local
features sharing problem as a vertex cover problem. 
When two or more inter-robot loop closure candidates
share a vertex in common, only the common vertex needs to be transmitted to
effectively compute all the associated relative pose measurements. Thus, by
computing the minimal vertex cover, optimally for bipartite graphs and
approximately with 3 robots or more, we obtain an exchange policy which avoids
redundant communication. 

\subsection{Inter-Robot Communication}

It is worth noting, that both for the spectral matching and the vertex exchange
policy, a temporary \textit{broker} needs to be dynamically elected among the
robots in communication range. The broker then computes the matches and sends
requests for the vertices to be transferred. In our current implementation, the
broker is simply the robot in range with the lowest ID according to our neighbor
management system, but it could be elected with a different decentralized
mechanism (e.g. based on the available computation resources onboard each
robot).

\section{Back-End}\label{sec:back_end}

The role of the back-end is to gather the odometry, intra- and inter-robot
loop closure measurements from the front-end in a pose graph, and then estimate
the most likely map and poses based on those noisy measurements.
As mentioned above, unlike other recent
systems~\cite{lajoiedoorslamdistributedonline2020,tianresourceawareapproachcollaborative2021}
based on distributed pose graph optimization, we opt for a simpler decentralized
approach. Similar to the front-end, a robot is dynamically elected to perform
the computation among the robots in communication range. The other robots share
their current pose graph estimates with the elected robot and receive the
updated estimates once the computation is completed. Importantly, any robot can
be temporarily elected through negotiation to perform the pose graph optimization during a
rendezvous between robots. \shortname performs the pose graph
optimization using the Graduated
Non-Convexity~\cite{yanggraduatednonconvexityrobust2020} solver, with the robust
Truncated Least Square loss.

To ensure convergence to a single global localization estimate after multiple
sporadic rendezvous without enforcing a central authority, we introduce an
anchor selection process to keep track of the current global reference frame.
During pose graph optimization, the anchor usually corresponds to a prior which
assigns a fixed value to the first pose of the graph. This anchor then becomes
the reference frame of the resulting estimate. In the beginning, all robots are
within their own local reference frames where the origin corresponds to their
first pose (i.e., initial position and orientation). Then, when some robots meet
for the first time (e.g. robots 0, 4 and 5), we choose the first pose of the
robot with the lowest ID (e.g. robot 0) as the anchor. Therefore, as a result of
the estimation process, the involved robots estimates share the same reference
frame (e.g. robot 0's first pose). In subsequent rendezvous (e.g. robots 2, 3
and 4), the anchor is selected based on the reference frame with the lowest ID
(e.g. robot 4's first pose is selected as the anchor since its reference frame
is robot 0's). After a few rendezvous, the robots converge to a single global
reference frame without requiring rendezvous including all robots (e.g. after
the second rendezvous, robots 2 and 3 are also within robot 0's reference
frame). This means that \shortname can scale to large groups of robots, through
iterative estimation among smaller groups of robots.

\section{Experimental Results}\label{sec:experiments}

\ExecuteMetaData[experimentation_figures.tex]{spectral-vs-greedy}

To evaluate the effectiveness of our proposed solutions for the ongoing
challenges in Collaborative Simultaneous Localization and Mapping, we conducted
extensive experiments on several public datasets, as well as in a real-world
deployment. Our experiments involved three robots exploring and mapping an
indoor environment and communicating via ad-hoc networking. We specifically
evaluated our key contributions to inter-robot loop closure detection and
decentralized C-SLAM estimation. Additionally, we present detailed statistics of
the communication and computation load during our real-world experiment,
providing insight into the system's performance and resource requirements.

\ExecuteMetaData[experimentation_figures.tex]{backend-table}

\subsection{Dataset Experiments}\label{subsec:data_exp}

We tested \shortname on seven sequences from
five different datasets. To demonstrate the flexibility of our framework, we
used IMUs, stereo cameras, lidars, or a combination as inputs. First, we tested
on the widely known autonomous driving KITTI~00 stereo
sequence~\cite{geigerareweready2012} which we split into two parts to simulate a
two-robots exploration. Second, we split the very large ($\sim$10km) KITTI360~09
lidar sequence~\cite{liaokitti360noveldataset2022} into 5 parts that contain a
large number of loop closures, making it particularly well suited for
inter-robot loop closure detection analysis. Third, we experimented on the first
three overlapping lidar sequences of the very recent GrAco
dataset~\cite{zhugracomultimodaldataset2023} acquired with custom ground
robots on a college campus. Fourth, we evaluate our system on the three lidar
Gate sequences of the M2DGR dataset~\cite{yinm2dgrmultisensormultiscenario2022}.
Fifth, we tested on three sequences of the recent C-SLAM-focused S3E
dataset~\cite{fengs3elargescalemultimodal2022}. To avoid tracking failures and
obtain more robust results on S3E sequences, we combined lidar-IMU odometry and
stereo camera-based inter-robot loop closure detection, highlighting the
versatility of \shortname. Overall, we chose the sequences with the most
trajectory overlaps to obtain more loop closures, and with available GPS ground
truth (except for S3E Laboratory). For simplicity and robustness, we used
off-the-shelf software~\cite{labbertabmapopensourcelidar2019} to
compute and provide the required odometry input to \shortname.
To better evaluate the inter-robot loop closure detection, we consider the
worst-case scenario in which the robots are within communication range only at
the end of their trajectories, such that they have to find loop closures between
their whole maps at once. This
scenario, analog to multiple robots exploring different parts of an environment
and meeting back at the end, is among the most challenging in terms of
communication and computation load, and therefore benefits the most from our
novel spectral candidate prioritization mechanism. We refer the reader to our
open-source implementation for all the parameters and configuration details of
the experiments.

\subsubsection{Inter-Robot Loop Closure Detection Evaluation}

In~\cref{fig:spectral-vs-greedy}, we compare the greedy and spectral inter-robot
loop closure candidate prioritization techniques with respect to algebraic
connectivity, and Absolute Translation Error (ATE). 
  Each approach is
used to prioritize the computation of loop closures from the same set of
candidates with a budget $B$ of 1, i.e. selecting one loop closure at a time. We plot each metric against the percentage of
loop closures computed within the set of candidate (x-axis). 
We perform prioritization successively up until all the possible matches are selected (i.e., 100\% of loop
closures computed). 
We expect that a better prioritization will acheive reasonable accuracy early on, with only a fraction of the matches selected. 
The ATE is computed 
against the final pose graph estimate containing all possible inter-robot loop
closures, and thus constitutes the best estimate we can achieve.
On the first row, we can see that, as intended, our spectral prioritization is
correctly maximizing the algebraic connectivity of the pose graph. 
On the second row, as expected, we can see that our spectral
prioritization decreases the error faster than the greedy prioritization. 
Overall, our experiments show that carefully selecting candidates early on requires the computation of fewer inter-robot loop closures to significantly reduce the estimation error (ATE).

\subsubsection{Decentralized C-SLAM Evaluation}

\ExecuteMetaData[experimentation_figures.tex]{rendezvous}

In~\cref{tab:backend}, we present the estimates computed in the back-end on all
the sequences for which GPS latitude and longitude data is available as ground
truth. We report the computation time on a AMD Ryzen 7 CPU and the total
communication required in kB. 
Using our same front-end, we compared our GNC-based decentralized back-end against two state-of-the-art distributed
approaches: the Distributed Gauss-Seidel (DGS) pose graph
optimization~\cite{choudharydistributedmappingprivacy2017} combined with
Pairwise Consistency Maximization
(PCM)~\cite{mangelsonpairwiseconsistentmeasurement2018} for outlier rejection as
used in~\cite{lajoiedoorslamdistributedonline2020}; and a distributed
implementation of Graduated Non-Convexity
(D-GNC)~\cite{tiankimeramultirobustdistributed2022} based on the RCBD
solver~\cite{tiandistributedcertifiablycorrect2021}. 
Our chosen back-end consistently achieved the highest level of accuracy (ATE)
whereas alternative methods occasionally fell short of generating reasonable estimates.
We also consistently outperforms the other approaches in terms of required communication and computation time.
Interestingly, when tested on KITTI-360 09, our dataset with the highest number of robots, both DGS+PCM and D-GNC take less computation time compared to GNC, yet they don't achieve equivalent accuracy and necessitate more than five times the amount of data transmission.
 While distributed
approaches benefit from the division of labour on large problems, more
research is required to obtain the same levels of accuracy, robustness, and
communication bandwidth. 
This justifies our practical choice of a simpler approach computing the back-end on a single decentrally-elected robot, which is more robust to communication failures and easier to implement.

\ExecuteMetaData[experimentation_figures.tex]{other-graphs}

In~\cref{fig:rendezvous}, we show the \shortname resulting estimates on the
KITTI360 09 sequence from four different rendezvous, defined as an
encounter in which a subset of robots are within communication range of each
other. Our anchor selection scheme ensures that by choosing the current
first pose estimate from the robot with the lowest reference frame ID (i.e.,
first poses of (a) robot 0, (b) robot 2, (c) robot 3), we can propagate the
global reference frame among the team of robots. In other words, we are able to
converge to a single global reference frame through successive estimations
between subsets of robots, without enforcing connectivity maintenance or a
central authority. This decentralized approach improves the scalability of the
system by relying only on local interactions among neighboring robots. We present
the \shortname solutions on the remaining dataset sequences
in~\cref{fig:other-graphs}.

\ExecuteMetaData[explanation_figures.tex]{real}

\subsection{Real-World Experiments}\label{subsec:real_exp}

To assess the viability of \shortname on resource-constrained platforms, we
deployed the system in an indoor parking lot and gathered statistics regarding
the computation time and communication load. As shown in~\cref{fig:top}, we
performed an online real-world demonstration with 3 different robots (Boston
Dynamics Spot, Agilex Scout, and Agilex Scout Mini), all equipped with an NVIDIA
Jetson AGX Xavier onboard computer, an Intel Realsense D455 camera, an Ouster
lidar OS0-64, a VectorNav VN100 IMU, and a GL-iNet GL-S1300 OpenWrt gateway for
ad hoc networking. We used lidars and IMUs for odometry and the RGB-D cameras
for inter-robot loop closure detection. 

\ExecuteMetaData[experimentation_figures.tex]{real-exp}

As stated in~\cref{tab:real_exp}, our robots travelled a total of 475~meters
during the experiment and produced a total of 3103~keyframes that needed to be
matches and verified in the search for inter-robot loop closures. The process
resulted in 67~loop closures, including 10 that were rejected by the GNC
optimizer. This large number of outliers is
attributable to the many similar-looking sections of the parking lot. 
\shortname acheived accurate localization with the transmission of only 
94.95~MB of data between the robots, excluding the visualization. The
communication load is mostly attributable to the front-end and thus dependent on
the number of keyframes. In~\cref{tab:real_exp}, we also report the average
sparsification and pose graph optimization times. We can observe that the
sparsification time, while being non-negligible, is lower than the pose graph
optimization. 
To mitigate this, we implemented sparsification and optimization within
separate threads. 
\section{Conclusions And Future Work}\label{sec:conclusion}

In this paper, we presented Swarm-SLAM, a comprehensive resource-efficient C-SLAM
framework that is designed to comply with essential properties of swarm
robotics. 
In future work, we 
 aim to investigate collaborative
domain calibration and/or uncertainty estimation in place
recognition~\cite{lajoieselfsup2022} to reduce the prevalence of measurement
outliers among inter-robot loop closures, and therefore increase the overall
accuracy and resilience of C-SLAM.
Overall, we hope that our open-source framework will be useful as a testbed for
the research and development of new methods and techniques in place recognition,
inter-robot loop closure detection, multi-robot pose graph optimization, and
other open-problems.

\bibliographystyle{IEEEtran}

\end{document}